\documentclass{article}

\PassOptionsToPackage{numbers, compress}{natbib}

\usepackage[preprint]{neurips_2026}

\usepackage[utf8]{inputenc} % allow utf-8 input
\usepackage[T1]{fontenc}    % use 8-bit T1 fonts
\usepackage{hyperref}       % hyperlinks
\usepackage{url}            % simple URL typesetting
\usepackage{booktabs}       % professional-quality tables
\usepackage{makecell}       % multiline table headers
\usepackage{amsmath}        % math typesetting (cases, align, etc.)
\usepackage{amsfonts}       % blackboard math symbols
\usepackage{nicefrac}       % compact symbols for 1/2, etc.
\usepackage{microtype}      % microtypography
\usepackage[table]{xcolor}  % colors + \arrayrulecolor for tables
\usepackage{graphicx}       % figures
\usepackage{wrapfig}        % figures wrapped by text
\usepackage{pifont}         % \ding{N} dingbats (e.g., \ding{182} = circled 1)
\let\oldding\ding% Store old \ding in \oldding
\renewcommand{\ding}[2][1]{\scalebox{#1}{\oldding{#2}}}

\title{Inline Critic Steers Image Editing}

\author{%
  \textbf{Weitai Kang}$^{1}${ }
  \textbf{Xiaohang Zhan}$^{2}${ }
  \textbf{Yizhou Wang}$^{2}${ }
  \textbf{Mang Tik Chiu}$^{2}${ }
  \textbf{Jason Kuen}$^{2}${ } \\
  \textbf{Kangning Liu}$^{2}${ }
  \textbf{Yan Yan}$^{1}${ } \\[0.4em]
  $^1$University of Illinois Chicago{ }
  $^2$Adobe
}

\begin{document}

\maketitle

\begin{abstract}
    Instruction-based image editing exhibits heterogeneous difficulty not only across cases but also across regions of an image, motivating refinement approaches that allocate correction to where the model struggles. Existing refinement signals arrive late, after a fully generated image or a completed denoising step. We ask whether such a signal can act \emph{within} an ongoing forward pass. To investigate this, we probe a frozen image-editing model and find that although generation capability emerges only in the last few layers, the error pattern is already set in early layers (rank correlation $\rho=0.83$ with the final-layer error map). Based on this, we introduce \textbf{Inline Critic}, a learnable token that critiques a frozen model's predictions at its intermediate layers and steers its hidden states to refine generation during the forward pass. A three-stage recipe is proposed to stabilize the training from learning how to critique to steering generation. As a result, we achieve state of the art on GEdit-Bench (7.89), a $+9.4$ gain on RISEBench over the same backbone, and the strongest open-source result on KRIS-Bench (81.92, surpassing GPT-4o). We further provide analyses showing that the critic genuinely shapes the model's attention and prediction updates at subsequent layers.
\end{abstract}
\section{Introduction}
\label{sec:intro}

% Instruction-based image editing aims to modify an input image according to a natural-language instruction. Despite recent progress, the task remains challenging because its difficulty is highly heterogeneous. Some cases demand broad yet precise modifications, while others require only a small change with most of the image left untouched~\citep{ye2025imgedit, liu2025step1xedit, luo2026editscore}. Even within a single image, different regions vary substantially in their difficulty~\citep{schusterbauer2026patchforcing}. Instructions also vary in complexity: some are direct descriptive edits, while others require heavy reasoning or world knowledge~\citep{wu2025krisbench, zhao2025risebench, yin2025reasonedit, li2026thinkrledit}. An editor should therefore allocate different amounts of effort across cases.

Instruction-based image editing aims to modify an input image according to a natural-language instruction. Despite recent progress, the task remains challenging because its difficulty is highly heterogeneous, both within a single image and across cases. 
Some demand major yet precise modification while others only need small changes or remain untouched~\citep{schusterbauer2026patchforcing, ye2025imgedit, wu2025krisbench, zhao2025risebench, liu2025step1xedit}. An editor should therefore allocate different amounts of effort across regions and cases.

% \begin{figure*}[h]
%   \centering
%   \includegraphics[width=0.97\textwidth]{figs/crop_figure1_motivation_layout.pdf}
%   \vspace{-9pt}
%   \caption{{Motivation for inline critic.} \textbf{Top:} On a frozen pretrained editor, we attach lightweight heads at nine intermediate blocks (L6--L54) to probe the generation outputs, with the original final output at L60.
%   \textbf{Middle:} Each head runs its own multi-step denoising. Usable generations only appear in the last few blocks.
%   \textbf{Bottom:} Based on the final output, we compute a per-token MSE map against the ground truth in each block. After rank-normalization and pooling, we measure the Spearman correlation between each intermediate map and the final map.
%   \textbf{Results:} Although generation capability only emerges in the last few blocks, the spatial error pattern is set far earlier. Even at L6, agreement with the final-layer map already reaches $\rho = 0.83$, and grows monotonically with depth. We therefore ask whether this prior can be exploited within the forward pass to let the model recognize and correct its weak regions in time.}
%   \label{fig:motivation}
%   \vspace{-17pt}
% \end{figure*}
\begin{figure*}[h]
  \centering
  \includegraphics[width=0.97\textwidth]{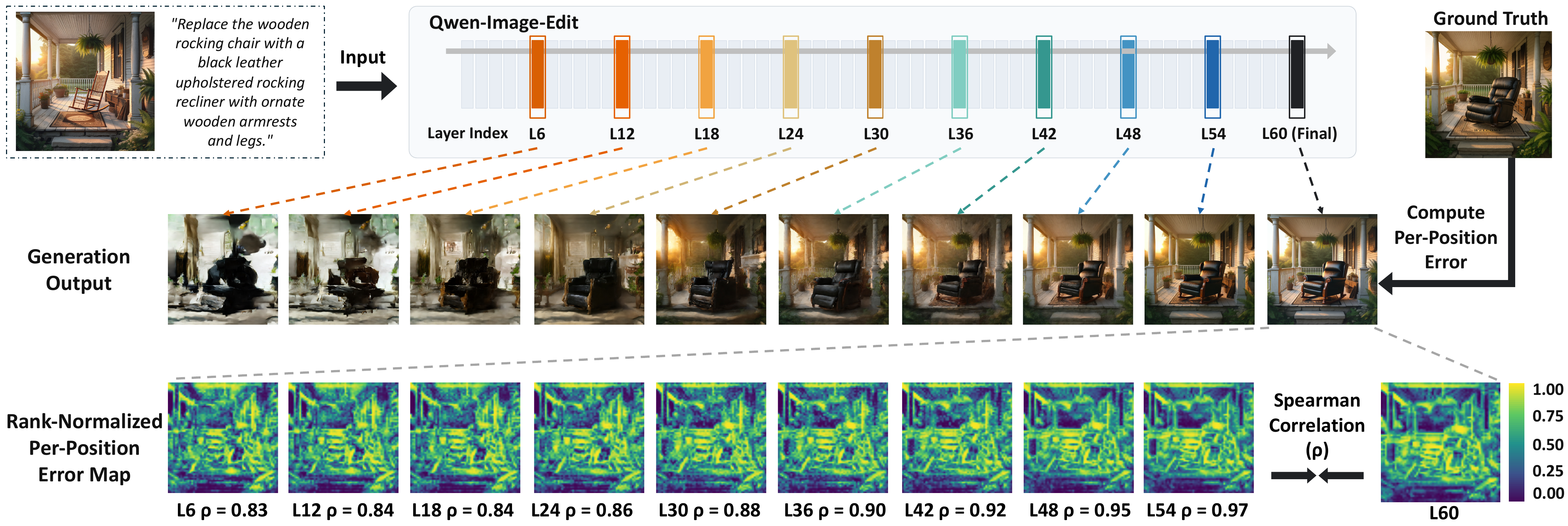}
  \caption{{Motivation for inline critic.} \textbf{Top:} On a frozen model, we add lightweight heads at intermediate blocks (L6--L54) to probe the generation outputs, with the original final output at L60.
  \textbf{Middle:} Each head runs its own multi-step denoising to generate an image. %Usable generations only appear in the last few blocks.
  \textbf{Bottom:} Based on the final output, we compute a MSE map against the ground truth in each block. After 8x pooling and rank-normalization, we measure the Spearman correlation between each intermediate map and the final map.
  \textbf{Results:} Although generation capability only emerges in the last few blocks, the spatial error pattern is set far earlier. Even at L6, agreement with the final-layer map already reaches $\rho = 0.83$, and grows monotonically with depth. We therefore ask whether this prior can be exploited within the forward pass to let the model recognize and correct its weak regions in time.}
  \label{fig:motivation}
\end{figure*}

Recent work addresses this issue by introducing \emph{refinement} signals that identify weak generations and allocate additional correction accordingly. One line of work computes this signal from a fully generated image: an external vision-language model or reward model critiques the output, which drives reranking, regeneration, or reinforcement learning~\citep{li2026thinkrledit, luo2026editscore, yin2025reasonedit, li2025uniworldv2, zhuo2025reflectionflow, li2025reflectdit}. Another line obtains the signal earlier in the generation process, from each completed denoising step: confidence, uncertainty, or difficulty estimates then adjust the current step or steer subsequent iterations~\citep{schusterbauer2026patchforcing, koulischer2025feedback, devita2025pixelwise, karras2024autoguidance, ahn2024pag, kou2024bayesdiff, epstein2023selfguidance, chefer2023attend, lezama2022tokencritic, chang2022maskgit}. Pushing this further, we are curious whether \emph{the refinement signal can act even earlier, within the ongoing forward pass}, to adjust the prediction before a step completes.

% To investigate this possibility, we analyze the intermediate hidden states of a strong pretrained image editor~\citep{wu2025qwenimage}. We freeze the model and train lightweight probe heads at multiple intermediate layers to assess their encoded generation ability. For each head's output, we compute a prediction error map against the ground truth, rank-normalize it, and compare it against the map of the final layer. As shown in Fig.~\ref{fig:motivation}, although intermediate-layer generation becomes competitive only near the last few layers, the relative error pattern is already stable within early layers, with Spearman's $\rho = 0.83$ against the final layer's map and even higher agreement in later layers. In other words, \emph{where the model performs well or poorly emerges long before the forward pass completes.}
To investigate this possibility, we analyze the intermediate hidden states of a strong pretrained image editor~\citep{wu2025qwenimage}. We freeze the model and train lightweight probe heads at multiple intermediate layers to assess their encoded generation ability. On 400 random samples drawn from different datasets, for each probe head's output we compute a prediction error map against the ground truth. We pool and rank-normalize the error maps, and compare them against the map of the final layer. As shown in Fig.~\ref{fig:motivation}, although intermediate-layer generation becomes competitive only near the last few layers, the relative error pattern is already stable within early layers, with a Spearman's $\rho = 0.83$ against the final layer's map averaged by steps and even higher agreement in later layers. In other words, \emph{where the model performs well or poorly emerges long before the forward pass completes.}

% Motivated by this observation, we leverage the early-emerging error pattern to introduce a refinement signal within the ongoing forward pass. We introduce \textbf{Inline Critic}: a single learnable token added to a frozen pretrained editor that critiques the model's prediction at multiple intermediate layers and steers their hidden states to refine the encoded generation behavior. 
% The method is trained via a three-stage recipe that decouples diagnostic learning from generation steering: 
% \raisebox{-1.1pt}{\ding[1.1]{182\relax}} 
% We train lightweight heads on top of detached hidden states at multiple intermediate layers with the original generation loss, probing the frozen encoded generation behavior without affecting the model.
% \raisebox{-1.1pt}{\ding[1.1]{183\relax}} 
% We append a learnable token (the critic token) to the input sequence, alongside the existing text and image tokens, and train it to predict (critique) the per-position generation error at each probed layer. Importantly, to develop the critique ability of the critic token stably, we mask it from being attended to by the other tokens, thus the model still remains unaffected and the critic's target (the generation error) does not drift.
% \raisebox{-1.1pt}{\ding[1.1]{184\relax}} 
% We unmask the critic token so the other tokens can attend to it, and jointly optimize the model's final generation objective with the critic objective, forcing the critic token to steer the latent representations to refine the encoded generation behavior at each layer.
Motivated by this, we aim to inline a critique into the forward pass, so that the model can learn to judge its emerged difficulty pattern. This critique is coupled with the generation objective, forcing it to be a useful refinement signal. 
Therefore, we introduce \textbf{Inline Critic}, a learnable token added to a frozen model that critiques its prediction at multiple intermediate layers and steers its hidden states to refine generation. We stabilize training through a three-stage recipe. We first probe the generation behavior of intermediate layers, the error of which the critic then learns to predict and later steers the model's prediction at each layer. At inference, the model does not need to compute the probe heads or critic predictions, with just an additional critic token in the input sequence.

Our contributions are as follows: (i)~\textbf{Analysis}: by probing the intermediate layers, we find that the spatial pattern of \emph{where} the model struggles is already stable in early layers, long before the forward pass completes; (ii)~\textbf{Method}: motivated by this, we propose \textbf{Inline Critic}, a learnable token that critiques the model's prediction at intermediate layers and steers their hidden states to refine generation, together with a three-stage curriculum that first establishes the critique signal, then activates generation steering; (iii)~\textbf{Results}: our method reaches state of the art on benchmarks (e.g. $7.89$ on GEdit-Bench). We further provide analyses showing that the critic influence where our model re-allocates attention at the next layer and where it updates its prediction in subsequent layers.

\section{Related Work}
\label{sec:related}

\subsection{Refinement Mechanisms for Generation}
\label{sec:related-refinement}

We organize refinement methods by when the refinement signal becomes available. \emph{Post-generation} methods derive feedback from completed images~\citep{li2026thinkrledit, luo2026editscore, yin2025reasonedit, li2025uniworldv2, zhuo2025reflectionflow, li2025reflectdit}, typically using an external vision-language model or reward model. \emph{Post-step} methods derive feedback after a completed denoising forward pass and use it to adjust the current step or steer subsequent iterations~\citep{schusterbauer2026patchforcing, koulischer2025feedback, devita2025pixelwise, karras2024autoguidance, ahn2024pag, kou2024bayesdiff, epstein2023selfguidance, chefer2023attend, lezama2022tokencritic, chang2022maskgit}. Our method instead produces a spatial critique within the ongoing forward pass and lets later layers consume it before the denoising step completes.

\paragraph{Post-Generation Refinement.}
Post-generation methods improve generation by evaluating completed generation outputs and using the verdict to drive reranking, regeneration, or reinforcement-learning fine-tuning. Reflect-DiT~\citep{li2025reflectdit} feeds a vision-language model's textual critique back as in-context guidance for another generation attempt. ReflectionFlow~\citep{zhuo2025reflectionflow} scales inference-time optimization for text-to-image diffusion using VLM-derived reflection data. ReasonEdit~\citep{yin2025reasonedit} frames editing as a multimodal-LLM thinking--editing--reflection loop. ThinkRL-Edit~\citep{li2026thinkrledit}, UniWorld-V2~\citep{li2025uniworldv2}, and EditScore~\citep{luo2026editscore} optimize image editors with reinforcement learning signals from explicit reward models or implicit MLLM feedback. These methods are powerful for semantic assessment, but their feedback is only available after the image has been generated.

\paragraph{Post-Step Refinement.}
Post-step methods use signals extracted after a denoising forward pass to adjust the current step or subsequent iterations. Diffusion Self-Guidance~\citep{epstein2023selfguidance} extracts internal features and backpropagates a property-specific objective to steer sampling. Attend-and-Excite~\citep{chefer2023attend} similarly extracts attention activations and backpropagates an objective on them to update the latent. Perturbed-Attention Guidance~\citep{ahn2024pag} and Autoguidance~\citep{karras2024autoguidance} construct a degraded or weak prediction at each step and use the difference as guidance, while Feedback Guidance~\citep{koulischer2025feedback} formulates a state-dependent per-step feedback signal. Difficulty- and uncertainty-aware samplers provide finer spatial signals: Patch Forcing~\citep{schusterbauer2026patchforcing} estimates per-patch difficulty to allocate more denoising effort to hard regions, and BayesDiff~\citep{kou2024bayesdiff} and \citet{devita2025pixelwise} estimate pixel-wise uncertainty for diffusion sampling. In discrete masked generation, Token-Critic~\citep{lezama2022tokencritic} scores generated visual tokens for remasking, and MaskGIT~\citep{chang2022maskgit} uses token confidence to schedule iterative decoding. These methods show that local difficulty and uncertainty are useful refinement signals, but the signal is produced only after the forward pass completes, rather than emerging within the ongoing forward pass.

\subsection{Learnable Tokens for Generation}
\label{sec:related-tokens}

Learnable tokens provide a parameter-efficient way to adapt frozen generators. Textual Inversion~\citep{gal2023textualinversion} represents a new visual concept as a learned token embedding inserted into the prompt sequence. DreamArtist~\citep{dong2022dreamartist} learns positive and negative prompt embeddings, while ReVersion~\citep{huang2024reversion} learns tokens that encode relations between subjects. Several works make the learned representation layer- or time-dependent: NeTI~\citep{alaluf2023neti} parameterizes the token as a function of layer index and denoising timestep, P+~\citep{voynov2023pplus} uses different embeddings across cross-attention layers, and ProSpect~\citep{zhang2023prospect} assigns learnable tokens to different denoising stages. These tokens primarily act as conditioning codes that inject new concepts, relations, or attributes. Our critic token serves a different role: it is trained to critique the model by predicting spatial generation error from intermediate hidden states, and to expose this critique back to later layers to steer the generation behavior.

\section{Method}
\label{sec:method}

\begin{figure}[t]
  \centering
  \includegraphics[width=\textwidth]{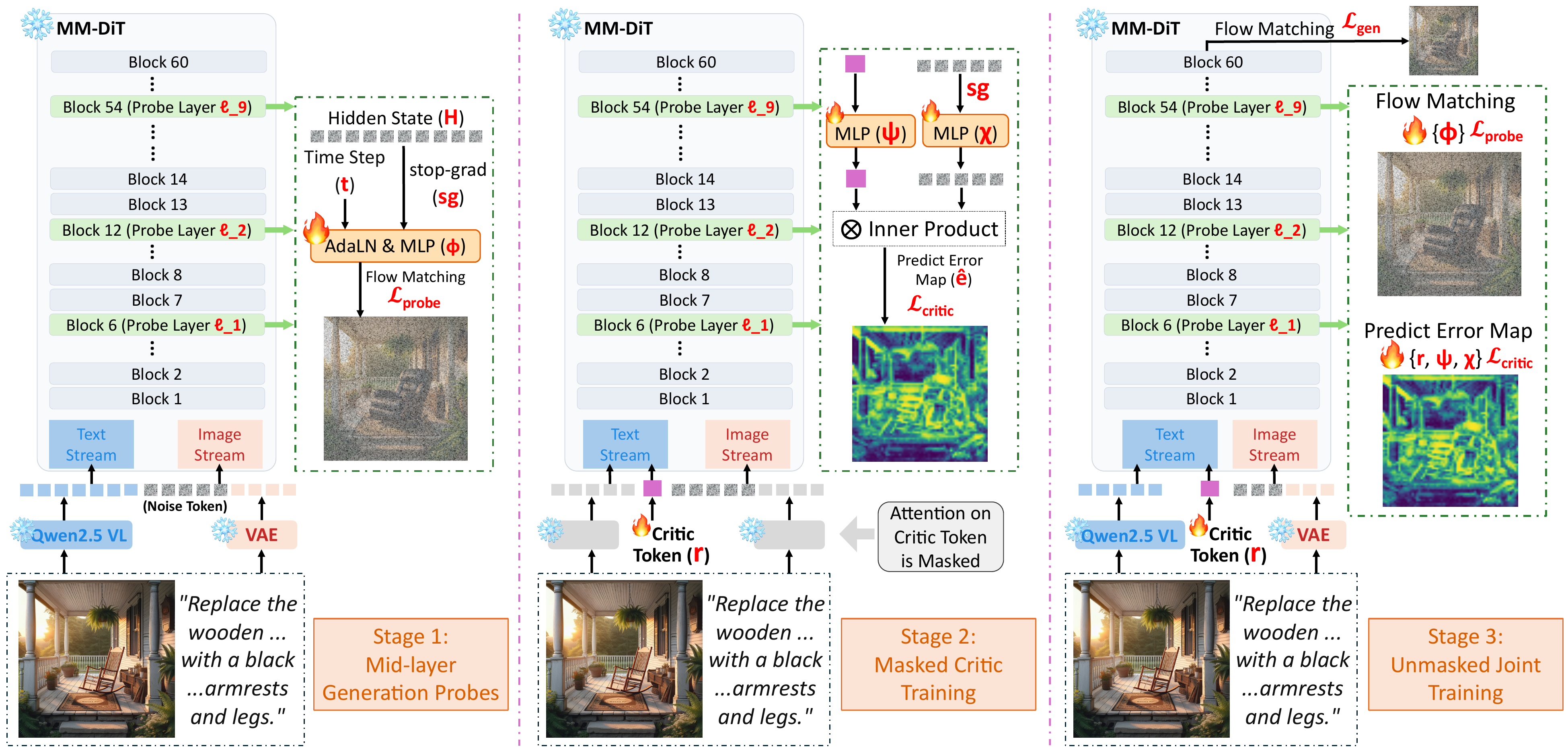}
  \caption{\textbf{Three-stage training of Inline Critic.} \textbf{Left}: Stage 1 trains probe heads at several intermediate layers to predict the generation target from frozen hidden states. \textbf{Middle}: Stage 2 adds a critic token that is trained, at every probed block, to predict the error of Stage-1's probes; an attention mask hides it from other tokens so the model and the probes are unaffected. \textbf{Right}: Stage 3 removes the mask and trains the critic jointly with the generation loss to steer generation.}
  \label{fig:method}
\end{figure}

We first provide a brief review of the image editing backbone (Sec.~\ref{sec:method-prelim}), then introduce our proposed Inline Critic (Sec.~\ref{sec:method-overview}) and its three-stage training recipe (Sec.~\ref{sec:method-stage1}--\ref{sec:method-stage3}), illustrated in Fig.~\ref{fig:method}.

\subsection{Preliminary: Qwen-Image-Edit}
\label{sec:method-prelim}

Our backbone is the open-source instruction-based image editor Qwen-Image-Edit~\citep{wu2025qwenimage}. A Qwen2.5-VL~\citep{bai2025qwen25vl} encodes the editing instruction together with the source image into conditioning tokens $\mathbf{c}$, forming the \emph{text stream}. The \emph{image stream} consists of $S$ \emph{noise tokens} that the model denoises, concatenated with source image tokens encoded by a Wan2.1-VAE~\citep{wan2025}. A double-stream diffusion transformer of $L = 60$ MM-DiT blocks~\citep{esser2024sd3} operates jointly on the two streams: at each block, the text-stream and image-stream tokens are concatenated along the sequence dimension and mixed by a joint self-attention, and modality-specific feed-forward and adaptive-LayerNorm layers update each stream. After block $L$, the noise-token outputs form the per-position velocity prediction.

Training follows the rectified-flow objective~\citep{liu2023flow, lipman2023flow}. Let $\mathbf{x}_0 \in \mathbb{R}^{S \times d_z}$ be the VAE-encoded target latent and $\mathbf{x}_1 \sim \mathcal{N}(\mathbf{0}, \mathbf{I})$ a noise sample. Given a time $t \in [0, 1]$, noise tokens are set to $\mathbf{x}_t = (1-t)\,\mathbf{x}_0 + t\,\mathbf{x}_1$, and the network's per-position velocity prediction $f_\theta(\mathbf{x}_t, t, \mathbf{c})$ is regressed against the target velocity $\mathbf{v}^\star = \mathbf{x}_1 - \mathbf{x}_0$ with the mean-squared-error (MSE) loss
\begin{equation}
\mathcal{L}_{\text{gen}}(\theta) \;=\; \mathbb{E}_{t,\, \mathbf{x}_0,\, \mathbf{x}_1,\, \mathbf{c}}\!\left[\big\| f_\theta(\mathbf{x}_t, t, \mathbf{c}) - \mathbf{v}^\star \big\|^2 \right].
\label{eq:gen-loss}
\end{equation}

\subsection{Inline Critic}
\label{sec:method-overview}

Inline Critic adds two lightweight modules to the frozen editor. First, we pick $K$ intermediate blocks $\mathcal{L}^* = \{\ell_1, \dots, \ell_K\} \subseteq \{1, \dots, L\}$ uniformly across the network. At each $\ell \in \mathcal{L}^*$, a probe head $g_\phi^{(\ell)}$ takes the detached noise token hidden states $\mathbf{H}^{(\ell)} \in \mathbb{R}^{S \times d}$ as input and is supervised by the same rectified-flow target $\mathbf{v}^\star$ as the final layer; its per-position prediction error against $\mathbf{v}^\star$ forms the error map that the critic later learns to predict. Second, we append a learnable critic token $\mathbf{r} \in \mathbb{R}^d$ to the text stream, giving conditioning $[\mathbf{c}; \mathbf{r}]$. At every probed layer $\ell$, small projection layers turn its hidden state $\mathbf{r}^{(\ell)}$ into a per-position estimate of this error map, and the critic token is later trained to influence the model's prediction through attention. The probes and projection layers participate only in training; at inference, they are dropped and only the critic token remains active in the forward pass.

Co-training all components from initialization would make the critic learn against a target that changes as the critic begins to influence the backbone, an instability analogous to the moving-target issue addressed by target networks in deep Q-learning~\citep{mnih2015dqn}. We therefore stage the training so that the critic learns against a stable target before being released to influence the model. We detail each stage in the following subsections.

\subsubsection{Stage 1: Mid-layer Generation Probes}
\label{sec:method-stage1}

As illustrated in Fig.~\ref{fig:method} (left), we instantiate each probe $g_\phi^{(\ell)}$ as
\begin{equation}
g_\phi^{(\ell)}(\mathbf{H}^{(\ell)}, t) \;=\; \mathrm{MLP}_\phi^{(\ell)}\!\Big( \mathrm{AdaLN}_\phi^{(\ell)}\!\big(\,\mathrm{sg}(\mathbf{H}^{(\ell)}),\, \tau(t)\big) \Big),
\label{eq:probe-arch}
\end{equation}
where $\mathrm{sg}(\cdot)$ is the stop-gradient operator, $\tau(t)$ is the time embedding shared with the backbone, $\mathrm{AdaLN}_\phi^{(\ell)}$ is a continuous adaptive LayerNorm conditioned on $\tau(t)$ (the same module type used inside the MM-DiT blocks), and $\mathrm{MLP}_\phi^{(\ell)}$ is a 3-layer feed-forward head whose output matches the per-position velocity. Only $\phi$ is trainable in this stage. The probes are trained with the original rectified-flow loss of Eq.~\ref{eq:gen-loss} applied at every probed layer:
\begin{equation}
\mathcal{L}_{\text{probe}}(\phi) \;=\; \frac{1}{K} \sum_{\ell \in \mathcal{L}^*} \mathbb{E}_{t,\, \mathbf{x}_0,\, \mathbf{x}_1,\, \mathbf{c}}\!\left[\big\| g_\phi^{(\ell)}(\mathbf{H}^{(\ell)}, t) - \mathbf{v}^\star \big\|^2\right].
\label{eq:probe-loss}
\end{equation}

\subsubsection{Stage 2: Masked Critic Training}
\label{sec:method-stage2}

As illustrated in Fig.~\ref{fig:method} (middle), we now introduce the critic token $\mathbf{r}$ together with $K$ pairs of small projection heads $\{\psi^{(\ell)}, \chi^{(\ell)}\}_{\ell \in \mathcal{L}^*}$ (each a 3-layer MLP). At every probed layer $\ell$ and noise token position $i$, the critic prediction is the inner product
\begin{equation}
\hat{e}^{(\ell)}_i \;=\; \big\langle\, \psi^{(\ell)}\!\big(\mathbf{r}^{(\ell)}\big),\; \chi^{(\ell)}\!\big(\mathrm{sg}(\mathbf{H}^{(\ell)}_i)\big) \,\big\rangle, \qquad i = 1, \dots, S,
\label{eq:critic-pred}
\end{equation}
where $\mathbf{H}^{(\ell)}_i \in \mathbb{R}^d$ is the hidden state of the $i$-th noise token at block $\ell$. The per-layer prediction error map is $\hat{\mathbf{e}}^{(\ell)} = (\hat{e}^{(\ell)}_1, \dots, \hat{e}^{(\ell)}_S)$.

To train the critic without changing the hidden states, we hide the critic token from the other tokens' attention at every block. Otherwise, the attention on the critic token would shift the noise tokens' hidden states and the probe error along with them, leaving the critic chasing a moving target. Letting $r$ denote the critic's index in the joint sequence, we add a mask $M$ to the pre-softmax scores with
\begin{equation}
M_{ij} \;=\; \begin{cases} -\infty & \text{if } i \neq r \text{ and } j = r, \\ 0 & \text{otherwise,}\end{cases}
\label{eq:iso-mask}
\end{equation}
where $i$ (query) and $j$ (key) range over all tokens in the joint sequence. The joint self-attention output for every non-critic token is then identical to a run with no critic token, so the model and the per-position generation error stay fixed. The critic itself attends to the other tokens normally.

The critic target is the frozen Stage-1 probes' per-position generation error. Since this error spans several orders of magnitude across noise tokens and time steps, we rescale it with $\log(1 + \cdot)$ to keep the learning from being dominated by a few high-value positions:
\begin{equation}
e^{(\ell)}_i \;=\; \log\!\Big(1 + \big\| g_\phi^{(\ell)}(\mathbf{H}^{(\ell)}, t)_i - \mathbf{v}^\star_i \big\|^2 \Big),
\label{eq:critic-target}
\end{equation}
where $g_\phi^{(\ell)}(\mathbf{H}^{(\ell)}, t)_i$ and $\mathbf{v}^\star_i$ are the probe output and the target velocity, and the gradient is stopped on $e^{(\ell)}_i$. The critic is trained by MSE, averaged over the $K$ probed layers and $S$ noise-token positions:
\begin{equation}
\mathcal{L}_{\text{critic}}(\mathbf{r}, \psi, \chi) \;=\; \frac{1}{K\, S} \sum_{\ell \in \mathcal{L}^*} \sum_{i=1}^{S} \Big( \hat{e}^{(\ell)}_i \,-\, \mathrm{sg}\!\big(e^{(\ell)}_i\big) \Big)^{\!2}.
\label{eq:critic-loss}
\end{equation}
Only $\{\mathbf{r}, \psi^{(\ell)}, \chi^{(\ell)}\}$ are trainable, while $\theta$ and $\phi$ remain frozen. In Stage 2, the critic predicts the per-position generation error at every probed layer without affecting the model or its own target.

\subsubsection{Stage 3: Unmasked Joint Training}
\label{sec:method-stage3}

As illustrated in Fig.~\ref{fig:method} (right), in Stage 3 we unmask the critic and jointly optimize the generation loss with the critic loss. This unmasking relaxes the strict target stability of Stage 2 in exchange for joint adaptation, which is acceptable since the critic has been pre-calibrated in Stage 2.
The joint loss is
\begin{equation}
\mathcal{L}_{\text{stage 3}} \;=\; \mathcal{L}_{\text{gen}} \;+\; \lambda_c\, \mathcal{L}_{\text{critic}} \;+\; \lambda_p\, \mathcal{L}_{\text{probe}},
\label{eq:stage3}
\end{equation}
where $\mathcal{L}_{\text{gen}}, \mathcal{L}_{\text{critic}}, \mathcal{L}_{\text{probe}}$ are Eqs.~\ref{eq:gen-loss}, \ref{eq:critic-loss}, \ref{eq:probe-loss}, and $\lambda_c, \lambda_p > 0$ are loss weights. The trainable parameters are $\{\phi, \mathbf{r}, \psi^{(\ell)}, \chi^{(\ell)}\}$, while $\theta$ remains frozen. Because the unmasked attention and $\mathcal{L}_{\text{gen}}$ now shift the noise token hidden states at intermediate blocks and thereby change the encoded generation behavior that the probes were trained against, we unfreeze $\phi$ and keep $\mathcal{L}_{\text{probe}}$ in the loss so that the probes can track this change. With the backbone frozen, unmasking enables the critic token $\mathbf{r}$ to steer intermediate hidden states through attention. Gradients from $\mathcal{L}_{\text{gen}}$ then flow into $\mathbf{r}$ via these attention connections, driving it to refine the model's generation behavior, while $\mathcal{L}_{\text{critic}}$ keeps the critic anchored to the updated probed generation error. In effect, this turns the critic token into an inline verifier that self-evaluates the intermediate generation at every probed layer and emits corrective feedback through attention, without any additional forward pass or external reward model. At inference, the critic token stays unmasked in the forward pass, and we no longer compute the probe outputs or the critic prediction. The per-step cost is therefore essentially identical to the original editor backbone.
\section{Experiments}
\label{sec:exp}

\subsection{Implementation}
\label{sec:exp-impl}

% \paragraph{Training.}
% We build Inline Critic on two public Qwen-Image-Edit~\citep{wu2025qwenimage} backbones, Qwen-Image-Edit-2509 and Qwen-Image-Edit-2511, yielding two variants that we refer to as Critic-v1 and Critic-v2. We use $K=9$ probes at $\mathcal{L}^* = \{6, 12, 18, 24, 30, 36, 42, 48, 54\}$, with $256$-dim outputs for $\psi^{(\ell)}$ and $\chi^{(\ell)}$. The training set is a $2.2$M-pair mixture from OmniEdit~\citep{wei2025omniedit}, UltraEdit~\citep{zhao2024ultraedit}, HQ-Edit~\citep{hui2025hqedit}, ShareGPT-4o-Image~\citep{chen2025sharegpt4oimage}, OpenGPT-4o-Image~\citep{wind2025opengpt4o}, UniREdit-Data-100K~\citep{maplebb2025uniredit}, Nano-Consistent-150K~\citep{yejy2025nanoconsistent}, and the SFT split of Pico-Banana-400K~\citep{apple2025picobanana}, processed at $512\!\times\!512$. For classifier-free guidance, the instruction is dropped with probability $0.05$, and with probability $0.5$ replaced by an offline Qwen3.5-27B~\citep{qwen2026qwen35} rewrite using the ThinkRL-Edit~\citep{li2026thinkrledit} prompt. Each stage runs for $4$ epochs with AdamW (lr $1\!\times\!10^{-4}$, $\beta=(0.9,0.999)$, weight decay $0.01$) and a cosine-with-restarts schedule (min-LR ratio $0.1$, $500$ warmup steps). The per-device batch is $4$ with gradient accumulation $4$, on $4$ nodes of $8$ A100-80G GPUs. Stage~3 uses $\lambda_c=\lambda_p=1$ in Eq.~\ref{eq:stage3}.
% \paragraph{Training.}

We train Inline Critic on two Qwen-Image-Edit~\citep{wu2025qwenimage} backbones, Qwen-Image-Edit-2509 and Qwen-Image-Edit-2511, resulting in Critic-v1 and Critic-v2, respectively. The model uses $K=9$ probe layers. Training is conducted on a $2.2$M-pair mixture of public image-editing datasets~\citep{wei2025omniedit,zhao2024ultraedit,hui2025hqedit,chen2025sharegpt4oimage,wind2025opengpt4o,maplebb2025uniredit,yejy2025nanoconsistent,apple2025picobanana}. The instruction is replaced by an Qwen3.5-27B~\citep{qwen2026qwen35} rewrite using the ThinkRL-Edit~\citep{li2026thinkrledit} prompt with probability $0.5$. All stages are trained for $4$ epochs with AdamW on $32$ A100-80G GPUs. 
We evaluate on RISEBench~\citep{zhao2025risebench}, KRIS-Bench~\citep{wu2025krisbench}, GEdit-Bench~\citep{liu2025step1xedit}, and ImgEdit~\citep{ye2025imgedit}, with the same rewrite applied on RISEBench and KRIS-Bench. We use $40$ denoising steps at guidance scale $4.0$. Details of dataset, augmentation, optimization, and training hyperparameters are provided in Appendix~\ref{appendix:training_details}. Qualitative results and limitations are put in Appendix~\ref{app:qualitative} and~\ref{app:limitations}.

\subsection{Main Results}
\label{sec:exp-main}

\begin{table}[t]
  \centering
  \caption{\textbf{Main results on GEdit-Bench~\citep{liu2025step1xedit} (EN split).} The best Overall score is in bold.}
  \vspace{-6pt}
  \label{tab:gedit}
  \small
  \begin{tabular}{lccc}
    \toprule
    Model & \makecell{Semantic\\Consistency} & \makecell{Perceptual\\Quality} & Overall \\
    \midrule
    % InstructPix2Pix~\citep{brooks2023instructpix2pix}                  & 3.58 & 5.49 & 3.68 \\
    % AnyEdit~\citep{yu2025anyedit}                                      & 3.18 & 5.82 & 3.21 \\
    % MagicBrush~\citep{zhang2023magicbrush}                             & 4.68 & 5.66 & 4.52 \\
    % UniWorld-V1~\citep{lin2025uniworldv1}                              & 4.93 & 7.43 & 4.85 \\
    % OmniGen~\citep{xiao2025omnigen}                                    & 5.96 & 5.89 & 5.06 \\
    % FLUX.1-Kontext [Dev]~\citep{blackforestlabs2025fluxkontext}        & 6.52 & 7.38 & 6.00 \\
    % OmniGen2~\citep{wu2025omnigen2}                                    & 7.16 & 6.77 & 6.41 \\
    % Gemini 2.0\textsuperscript{\dag}~\citep{google2025gemini2flash}    & 6.73 & 6.61 & 6.32 \\
    BAGEL~\citep{deng2025bagel}                                        & 7.36 & 6.83 & 6.52 \\
    FLUX.1-Kontext [Pro]~\citep{blackforestlabs2025fluxkontext}        & 7.02 & 7.60 & 6.56 \\
    Step1X-Edit~\citep{liu2025step1xedit}                              & 7.66 & 7.35 & 6.97 \\
    % UniPic2~\citep{skywork2025unipic2}                                 & --   & --   & 7.10 \\
    % GPT-Image-1 [High]\textsuperscript{\dag}~\citep{openai2025gptimage1} & 7.85 & 7.62 & 7.53 \\
    Qwen-Image-Edit-2509~\citep{wu2025qwenimage}                       & 8.15 & 7.86 & 7.54 \\
    UniWorld-Qwen-Image-Edit-2509~\citep{li2025uniworldv2}             & 8.36 & 7.87 & 7.76 \\
    ReasonEdit-Q~\citep{yin2025reasonedit}                             & 8.34 & 7.97 & 7.77 \\
    UniWorld-V2~\citep{li2025uniworldv2}                               & 8.39 & 8.02 & 7.83 \\
    \bf Critic-v1                                                        & 8.33 & 7.98 & 7.80 \\
    \bf Critic-v2                                                        & 8.48 & 8.02 & \textbf{7.89} \\
    \bottomrule
  \end{tabular}
\end{table}

\paragraph{GEdit-Bench.}
We start with general editing instructions on GEdit-Bench, scored by GPT-4.1. As shown in Tab.~\ref{tab:gedit}, our Critic-v2 achieves state-of-the-art performance with an $\mathbf{7.89}$ Overall score, also obtaining the highest Semantic Consistency ($8.48$) and tying for the best Perceptual Quality ($8.02$). Our Critic-v1 reaches $7.80$, outperforming UniWorld-Qwen-Image-Edit-2509~\citep{li2025uniworldv2} ($7.76$) under the same backbone. Both Semantic Consistency and Perceptual Quality improve over the backbone.

\begin{table}[t]
  \centering
  \caption{\textbf{Main results on KRIS-Bench~\citep{wu2025krisbench}.} Closed-source proprietary models are marked with \textsuperscript{\dag}. The best Overall score is in bold.}
  \vspace{-6pt}
  \label{tab:krisbench}
  \scriptsize
  \setlength{\tabcolsep}{1pt}
  \begin{tabular}{lccccccccccc}
    \toprule
    Model & \makecell{Attribute\\Perception} & \makecell{Spatial\\Perception} & \makecell{Temporal\\Prediction} & Factual & \makecell{Social\\Science} & \makecell{Natural\\Science} & Conceptual & \makecell{Logical\\Reasoning} & \makecell{Instruction\\Decomposition} & Procedural & Overall \\
    \midrule
    BAGEL-Think~\citep{deng2025bagel}                                & 67.42 & 68.33 & 58.67 & 66.18 & 63.55 & 61.40 & 61.92 & 48.12 & 50.22 & 49.02 & 60.18 \\
    Doubao\textsuperscript{\dag}~\citep{bytedance2025doubao}         & 70.92 & 59.17 & 40.58 & 63.30 & 65.50 & 61.19 & 62.23 & 47.75 & 60.58 & 54.17 & 60.70 \\
    Gemini 2.0\textsuperscript{\dag}~\citep{google2025gemini2flash}  & 66.33 & 63.33 & 63.92 & 65.26 & 68.19 & 56.94 & 59.65 & 54.13 & 71.67 & 62.90 & 62.41 \\
    Uni-CoT~\citep{qin2025unicot}                                    & 72.76 & 72.87 & 67.10 & 71.85 & 70.81 & 66.00 & 67.16 & 53.43 & 73.93 & 63.68 & 68.00 \\
    \bf Critic-v1                                                      & 84.06 & 86.42 & 55.97 & 79.96 & 83.35 & 82.57 & 82.76 & 67.91 & 74.39 & 71.15 & 79.02 \\
    GPT-4o\textsuperscript{\dag}~\citep{openai2025gpt4o}             & 83.17 & 79.08 & 68.25 & 79.80 & 85.50 & 80.06 & 81.37 & 71.56 & 85.08 & 78.32 & 80.09 \\
    \bf Critic-v2                                                      & 85.76 & 89.50 & 58.44 & 82.09 & 84.40 & 84.91 & 84.79 & 77.46 & 75.94 & 76.70 & \textbf{81.92} \\
    \bottomrule
  \end{tabular}
  \vspace{-8pt}
\end{table}

\paragraph{KRIS-Bench.}
We next evaluate on reasoning-heavy benchmarks, where the difficulty arises primarily from language reasoning and world knowledge. This shifts the bottleneck onto the understanding encoder (Qwen2.5-VL~\citep{bai2025qwen25vl}) of our backbone, so we first rewrite the original instruction with the stronger Qwen3.5-27B to provide a clearer, more direct prompt. We provide ablation on this rewrite in Sec.~\ref{sec:exp-ablation}. Table~\ref{tab:krisbench} reports per-sub-axis scores on KRIS-Bench, scored by GPT-4o with the official prompts. Our Critic-v2 ranks first overall at $\mathbf{81.92}$, surpassing GPT-4o~\citep{openai2025gpt4o} ($80.09$) by $+1.83$ and the strongest open-source baseline Uni-CoT~\citep{qin2025unicot} ($68.00$) by $+13.92$. 
% Our Critic-v1 ranks third overall at $79.02$, surpassing every open-source baseline as well as every closed-source proprietary model except GPT-4o. We further lead on the Factual and Conceptual aggregates, as well as on Spatial Perception ($89.50$) and Natural Science ($84.91$).

\begin{table}[t]
  \centering
  \caption{\textbf{Main results on RISEBench~\citep{zhao2025risebench} (\%).} Closed-source proprietary models are marked with \textsuperscript{\dag}. The best Overall score among open-source editors is in bold.}
  \vspace{-6pt}
  \label{tab:risebench}
  \resizebox{\textwidth}{!}{%
  \begin{tabular}{lcccccccc}
    \toprule
    Method & \makecell{Instruction\\Reasoning} & \makecell{Appearance\\Consistency} & \makecell{Visual\\Plausibility} & \makecell{Temporal\\Reasoning} & \makecell{Causal\\Reasoning} & \makecell{Spatial\\Reasoning} & \makecell{Logical\\Reasoning} & Overall \\
    \midrule
    \multicolumn{9}{l}{\emph{Closed-source proprietary models}} \\
    % Gemini-2.0-Flash-pre\textsuperscript{\dag}~\citep{google2025gemini2flashpreview}  & 49.9 & 68.4 & 84.9 & 11.8 & 14.4 & 11.0 & 2.4  & 10.0 \\
    % Seedream-4.0\textsuperscript{\dag}~\citep{bytedance2025seedream4}                & 58.9 & 67.4 & 91.2 & 17.6 & 13.3 & 11.0 & 7.1  & 12.2 \\
    % Gemini-2.0-Flash-exp\textsuperscript{\dag}~\citep{google2025gemini2flash}        & 48.9 & 68.2 & 82.7 & 9.4  & 16.7 & 23.0 & 4.7  & 13.9 \\
    % GPT-Image-1-mini\textsuperscript{\dag}~\citep{openai2025gptimage1mini}            & 54.1 & 71.5 & 93.7 & 25.9 & 31.1 & 33.0 & 9.4  & 25.3 \\
    GPT-Image-1\textsuperscript{\dag}~\citep{openai2025gptimage1}                    & 62.8 & 80.2 & 94.9 & 36.5 & 34.4 & 37.0 & 10.6 & 30.0 \\
    Gemini-2.5-Flash-Image\textsuperscript{\dag}~\citep{google2025gemini25flashimage}& 61.2 & 86.0 & 91.3 & 29.4 & 48.9 & 37.0 & 18.8 & 33.9 \\
    Gemini-3-pro-image\textsuperscript{\dag}~\citep{google2025gemini3proimage}       & 77.0 & 85.5 & 94.4 & 43.5 & 63.3 & 48.0 & 37.6 & 48.3 \\
    GPT-Image-2\textsuperscript{\dag}~\citep{openai2026gptimage2}                    & 73.8 & 89.3 & 94.9 & 45.9 & 66.7 & 50.0 & 34.1 & 49.4 \\
    GPT-Image-1.5\textsuperscript{\dag}~\citep{openai2025gptimage15}                 & 69.7 & 92.5 & 94.9 & 57.6 & 62.2 & 62.0 & 21.2 & 51.4 \\
    \midrule
    \multicolumn{9}{l}{\emph{Open-source editors}} \\
    % FLUX.1-Canny~\citep{blackforestlabs2025fluxkontext}         & 20.2 & 13.1 & 77.5 & 0.0  & 0.0  & 0.0  & 0.0  & 0.0 \\
    % HiDream-Edit~\citep{hidream2025i1}                          & 30.3 & 12.6 & 74.9 & 0.0  & 0.0  & 0.0  & 0.0  & 0.0 \\
    % Emu2~\citep{sun2024emu2}                                    & 22.6 & 38.2 & 78.3 & 1.2  & 1.1  & 0.0  & 0.0  & 0.5 \\
    % OmniGen~\citep{xiao2025omnigen}                             & 22.0 & 32.6 & 55.3 & 1.2  & 1.1  & 0.0  & 1.2  & 0.8 \\
    % Step1X-Edit~\citep{liu2025step1xedit}                       & 25.1 & 41.5 & 73.5 & 0.0  & 2.2  & 2.0  & 3.5  & 1.9 \\
    % Ovis-U1~\citep{aidcai2025ovisu1}                            & 33.9 & 52.7 & 72.9 & 1.1  & 3.3  & 4.0  & 2.4  & 2.8 \\
    FLUX.1-Kontext-Dev~\citep{blackforestlabs2025fluxkontext}   & 26.0 & 71.6 & 85.2 & 2.3  & 5.5  & 13.0 & 1.2  & 5.8 \\
    % BAGEL~\citep{deng2025bagel}                                 & 36.5 & 53.5 & 73.0 & 2.4  & 5.6  & 14.0 & 1.2  & 6.1 \\
    Qwen-Image-Edit-2509~\citep{wu2025qwenimage}                & 37.2 & 66.4 & 86.9 & 4.7  & 11.1 & 17.0 & 2.4  & 9.2 \\
    BAGEL (w/ CoT)~\citep{deng2025bagel}                        & 45.9 & 73.8 & 80.1 & 5.9  & 17.8 & 21.0 & 1.2  & 11.9 \\
    Qwen-Image-Edit-2511~\citep{wu2025qwenimage}                & 49.9 & 71.0 & 91.5 & 21.2 & 18.9 & 31.0 & 4.7  & 19.4 \\
    \bf Critic-v1                                                 & 59.1 & 85.4 & 92.2 & 37.6 & 46.7 & 42.0 & 8.2  & 34.2 \\
    \bf Critic-v2                                                 & 63.7 & 88.5 & 90.7 & 35.3 & 47.8 & 50.0 & 15.3 & \textbf{37.8} \\
    \bottomrule
  \end{tabular}}
  \vspace{-8pt}
\end{table}

\paragraph{RISEBench.}
RISEBench (Tab.~\ref{tab:risebench}, scored by GPT-4.1 with the official prompts) targets the same reasoning-heavy regime, with categories that probe temporal, causal, spatial, and logical reasoning. Our Critic-v2 attains an Overall of $\mathbf{37.8}$, improving over its backbone ($19.4$) by $+18.4$ and establishing a new state of the art among open-source editors; it also outperforms several proprietary models, including Gemini-2.5-Flash-Image\textsuperscript{\dag} ($33.9$) and GPT-Image-1\textsuperscript{\dag} ($30.0$).

\begin{table}[t]
  \centering
  \caption{\textbf{Main results on ImgEdit~\citep{ye2025imgedit}.} Recent strong baselines are re-evaluated under a consistent protocol. The best Overall score is in bold.}
  \vspace{-6pt}
  \label{tab:imgedit}
  \resizebox{\textwidth}{!}{%
  \begin{tabular}{lccccccccc|c}
    \toprule
    Method & action & add & adjust & bkgd & compose & extract & remove & replace & style & Overall \\
    \midrule
    Qwen-Image-Edit-2509~\citep{wu2025qwenimage}    & 4.769 & 4.407 & 4.323 & 4.255 & 3.652 & 3.427 & 4.512 & 4.605 & 4.850 & 4.311 \\
    UniWorld-Qwen-Image-Edit-2509~\citep{li2025uniworldv2} & 4.806 & 4.407 & 4.330 & 4.255 & 3.449 & 3.766 & 4.535 & 4.609 & 4.883 & 4.338 \\
    Qwen-Image-Edit-2511~\citep{wu2025qwenimage}    & 4.639 & 4.425 & 4.287 & 4.191 & 3.855 & 3.877 & 4.519 & 4.587 & 4.860 & 4.360 \\
    \bf Critic-v1                                     & 4.824 & 4.533 & 4.351 & 4.316 & 3.739 & 3.578 & 4.469 & 4.609 & 4.837 & 4.362 \\
    \bf Critic-v2                                     & 4.806 & 4.481 & 4.394 & 4.248 & 3.899 & 3.889 & 4.539 & 4.558 & 4.863 & \textbf{4.408} \\
    \bottomrule
  \end{tabular}}
\end{table}

\paragraph{ImgEdit.}
We score ImgEdit by GPT-4.1 with official prompts. We find that run-to-run variance of this benchmark can exceed typical method-to-method gaps. Compounding this issue, most prior works do not release their exact scoring configurations (e.g., GPT temperature). We therefore re-evaluate recent strong baselines under the consistent protocol in Tab.~\ref{tab:imgedit}. Our Critic-v2 obtains the best Overall score at $\mathbf{4.408}$, and Inline Critic consistently improves over the corresponding backbone on most categories ($7$ of $9$ for Critic-v1, $8$ of $9$ for Critic-v2). On the same 2509 backbone, our Critic-v1 ($4.362$) further outperforms the UniWorld-Qwen-Image-Edit-2509 RL-finetune~\citep{li2025uniworldv2} ($4.338$).

\begin{figure}[t]
  \centering
  \includegraphics[width=\textwidth]{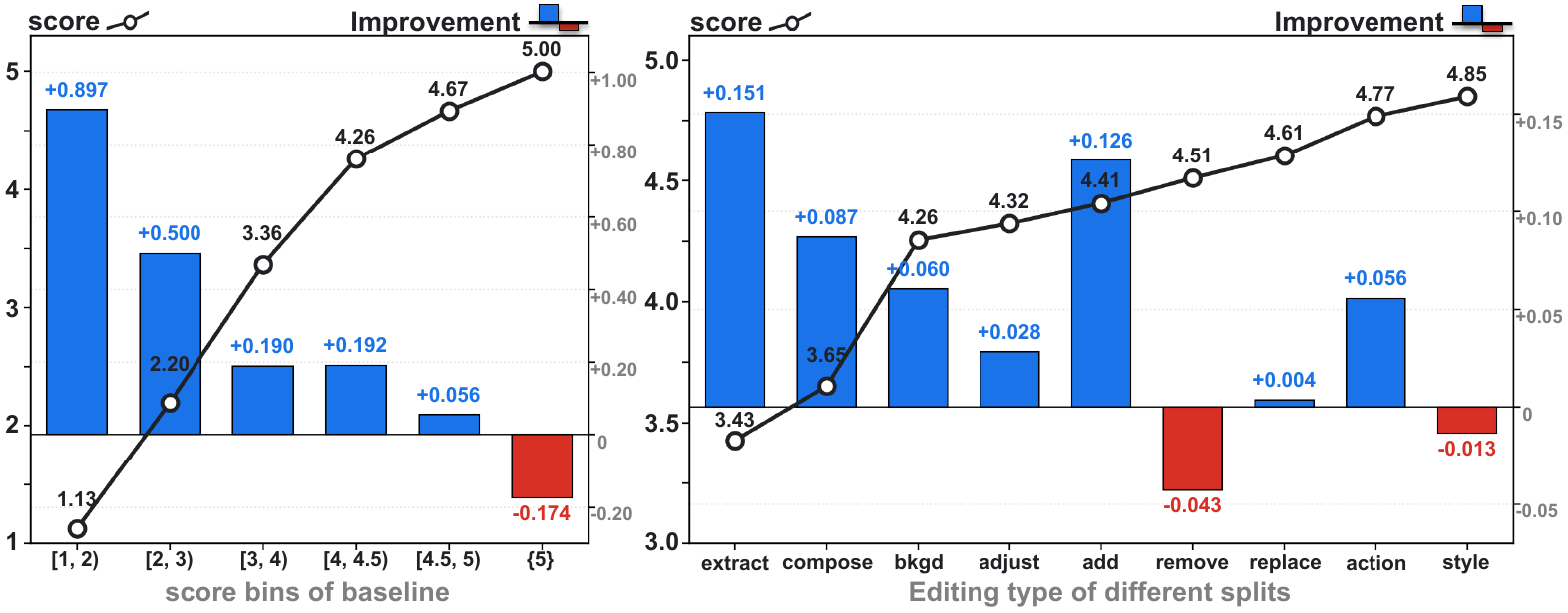}
  \vspace{-16pt}
  \caption{\textbf{Heterogeneity results on ImgEdit.} Bins are formed by baseline performance (left) and edit type (right). Our method's improvement concentrates on the hard cases.}
  \label{fig:long-tail}
\end{figure}

\paragraph{Heterogeneity results.}
Fig.~\ref{fig:long-tail} reports the performance in cases with different difficulty on ImgEdit with Qwen-Image-Edit-2509 as the baseline. The left panel groups samples into bins according to the baseline’s per-sample scores, while the right panel groups samples by edit type. Both panels show that the gain of Critic-v1 concentrates on the hard cases, with a bit of variance on the easy cases. Specifically, we bring $\mathbf{+0.897}$ on the $[1,2)$ score bin and $\mathbf{+0.151}$ on the hardest edit type \texttt{extract}.

\subsection{Ablation Studies}
\label{sec:exp-ablation}

In ablation study, we use 288K samples from the training set and train for one epoch on Qwen-Image-Edit-2509. For reproducibility, we choose Qwen3.5-27B as the judge at temperature $0$. 
% Our final model combines $K{=}9$ probed layers with rewrite augmentation, as reported in Sec.~\ref{sec:exp-main}.

% =========================== Table 1 ===========================
\begin{table}[t]
\caption{\textbf{Ablation of design choices on ImgEdit.} ``Learnable token'' adds a single learnable conditioning token trained with only the generation loss. $K$ denotes the number of probed layers. ``3-stage'' follows the training curriculum of Sec.~\ref{sec:method-overview}; ``single-stage'' uses only stage~3 of Sec.~\ref{sec:method-overview}. ``rewrite aug.'' replaces the training instruction with a Qwen3.5-27B rewrite~\citep{li2026thinkrledit} with probability $0.5$.}
\vspace{-6pt}
\label{tab:abl-imgedit}
\centering
\scriptsize
\setlength{\tabcolsep}{2.6pt}
\begin{tabular}{lcccccccccc}
\toprule
Method & action & add & adjust & bkgd & compose & extract & remove & replace & style & \textbf{Overall} \\
\midrule
Qwen-Image-Edit-2509 (backbone)                                  & $4.750$ & $4.179$ & $4.174$ & $3.961$ & $3.725$ & $3.365$ & $4.318$ & $4.268$ & $4.797$ & $\mathbf{4.171}$ \\
\arrayrulecolor{black!25}\midrule\arrayrulecolor{black}
\quad + LoRA                                        & $4.815$ & $4.088$ & $3.940$ & $4.060$ & $3.812$ & $3.444$ & $4.341$ & $4.304$ & $4.850$ & $\mathbf{4.184}$ \\
\quad + Learnable token          & $4.843$ & $4.214$ & $4.181$ & $4.184$ & $3.696$ & $3.439$ & $4.302$ & $4.362$ & $4.707$ & $\mathbf{4.214}$ \\
\arrayrulecolor{black!25}\midrule\arrayrulecolor{black}
\quad + Inline Critic, single-stage ($K{=}4$)               & $4.796$ & $4.109$ & $4.145$ & $4.028$ & $4.000$ & $3.470$ & $4.380$ & $4.199$ & $4.790$ & $\mathbf{4.213}$ \\
\quad + Inline Critic, 3-stage ($K{=}4$)                    & $4.861$ & $4.168$ & $4.241$ & $4.085$ & $3.884$ & $3.544$ & $4.399$ & $4.294$ & $4.793$ & $\mathbf{4.252}$ \\
\quad + Inline Critic, 3-stage ($K{=}9$)                    & $4.852$ & $4.249$ & $4.153$ & $4.234$ & $3.725$ & $3.521$ & $4.450$ & $4.286$ & $4.797$ & $\mathbf{4.252}$ \\
\quad + Inline Critic, 3-stage ($K{=}4$) + rewrite aug.     & $4.935$ & $4.197$ & $4.145$ & $4.128$ & $3.826$ & $3.715$ & $4.395$ & $4.297$ & $4.790$ & $\mathbf{4.270}$ \\
\bottomrule
\end{tabular}
\end{table}

As shown in Tab.~\ref{tab:abl-imgedit}, simply training with additional data by LoRA brings only a very limited improvement. For Inline Critic, directly using only the stage 3 of Sec.~\ref{sec:method-overview} is suboptimal. In that case, the critic target keeps moving before the critique training stabilizes, and the generation loss might pull the critic token toward acting as a generic learnable conditioning token. Consequently, single-stage Inline Critic reaches $4.213$, essentially matching the learnable token baseline at $4.214$. Instead, introducing our three-stage training recipe can improve Overall to $4.252$. We also find that the choice of $K$ is not sensitive: increasing from $K{=}4$ to $K{=}9$ gives the same Overall score (both at $4.252$). We adopt $K{=}9$ in the final model mainly to add a small amount of trainable capacity and to better expose the behavior of different model layers. Finally, rewrite augmentation provides a further improvement, raising Overall to $4.270$. On the reasoning-heavy RISEBench, Tab.~\ref{tab:abl-rewrite-rise} shows that Inline Critic consistently improves the backbone under both instruction settings: from $9.2$ to $18.6$ ($+9.4$) with raw instructions, and from $28.9$ to $34.2$ ($+5.3$) with rewritten instructions.

% =========================== Table 2 ===========================
\begin{table}[t]
\caption{\textbf{Ablation of test-time instruction rewrite on RISEBench~\citep{zhao2025risebench} (\%).} We compare Qwen-Image-Edit-2509 and Critic-v1 under raw and rewritten test instructions. The Qwen3.5-27B rewriter follows the ThinkRL-Edit~\citep{li2026thinkrledit} template. The best Overall score is in bold.}
\label{tab:abl-rewrite-rise}
\centering
\resizebox{\textwidth}{!}{%
\begin{tabular}{lcccccccc}
\toprule
Method & \makecell{Instruction\\Reasoning} & \makecell{Appearance\\Consistency} & \makecell{Visual\\Plausibility} & \makecell{Temporal\\Reasoning} & \makecell{Causal\\Reasoning} & \makecell{Spatial\\Reasoning} & \makecell{Logical\\Reasoning} & Overall \\
\midrule
\multicolumn{9}{l}{\emph{Raw instruction}} \\
Qwen-Image-Edit-2509~\citep{wu2025qwenimage} & $37.2$ & $66.4$ & $86.9$ & $4.7$ & $11.1$ & $17.0$ & $2.4$ & $9.2$ \\
Critic-v1 & $44.9$ & $73.2$ & $89.7$ & $7.1$ & $24.4$ & $32.0$ & $8.2$ & $18.6$ \\
\midrule
\multicolumn{9}{l}{\emph{Rewritten instruction}} \\
Qwen-Image-Edit-2509~\citep{wu2025qwenimage} & $57.1$ & $85.5$ & $91.5$ & $36.5$ & $32.2$ & $41.0$ & $3.5$ & $28.9$ \\
Critic-v1 & $59.1$ & $85.4$ & $92.2$ & $37.6$ & $46.7$ & $42.0$ & $8.2$ & $\mathbf{34.2}$ \\
\bottomrule
\end{tabular}}
\end{table}

\begin{figure*}[t]
\centering
\includegraphics[width=\textwidth]{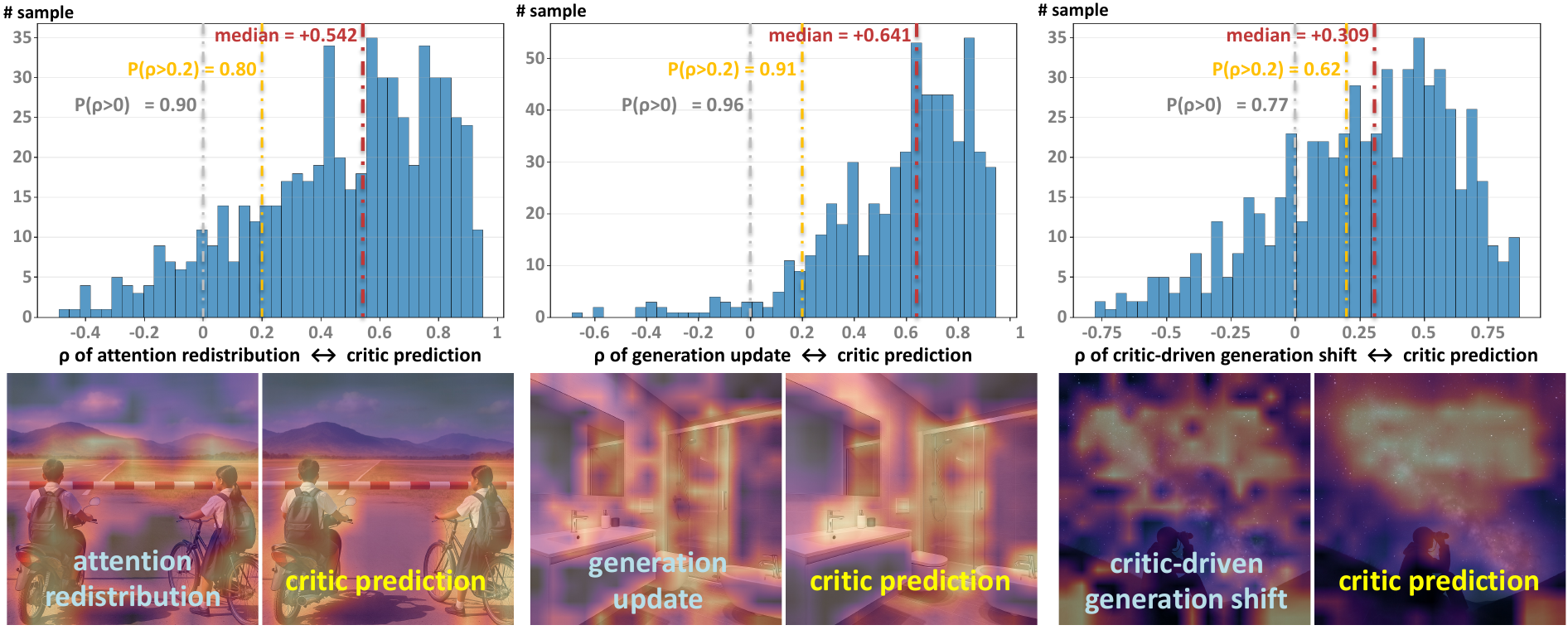}
\caption{\textbf{Top}: per-sample histogram of $\rho$ between the different model-side spatial map and the critic's prediction. \textbf{Bottom}: one sample showing the spatial map (left) and critic's prediction (right).}
\label{fig:analysis}
\end{figure*}

\subsection{Analysis}
\label{sec:exp-analysis}

We analyze whether the critic token truly steers spatial changes inside the backbone.
Our experiments show that three behaviors align with the critic prediction: A) the spatial redistribution of
the model's attention, B) the spatial update of the model's generation behavior, and C) the difference in
generation shift between runs with and without the critic token.

\paragraph{Setup.}
All analyses use $n=600$ held-out validation samples. We empirically inspect layer $L36$ to $L42$ at
denoising step $t=38$, where the critic is better calibrated, thereby reducing the confound that the critic might be
under-trained at the analyzed point. All Spearman correlations are computed in the same way as in
Fig.~\ref{fig:motivation}. Let $E=\hat{e}^{(36)}$ be the critic's predicted error map. For each experiment, we
define a spatial-change map $Y$ on the noise tokens and measure its agreement with $E$ by
\begin{equation}
\rho = \mathrm{Spearman}(Y, E).
\label{eq:analysis-rho}
\end{equation}

\paragraph{Experiment A: attention redistribution.}
We first measure where the model reallocates attention between $L36$ and $L37$, and compare this spatial pattern
against the critic's prediction $E$. For noise tokens acting as the key, $\alpha^{(\ell)}$ is a map about the sum of pre-softmax attention scores each token receives from all queries. We then use
\begin{equation}
Y_A = |\alpha^{(37)}-\alpha^{(36)}|.
\end{equation}
As shown in Fig.~\ref{fig:analysis} (left), the median $\rho$ is +0.542, with $P(\rho > 0)=0.90$ and $P(\rho > 0.2)=0.80$. The two maps are positively aligned: regions flagged as harder (i.e., with higher error prediction) by the critic also exhibit the higher attention changes, indicating that the critic redirects attention.

\paragraph{Experiment B: generation behavior update.}
We next measure where the model's encoded generation behavior changes between $L36$ and $L42$, and compare
this spatial pattern against the critic's prediction $E$. For each probe layer, $m^{(\ell)}$ is a map about
the error between the probe layer $\ell$ and the ground-truth. We then use
\begin{equation}
Y_B = |m^{(42)}-m^{(36)}|.
\end{equation}
Fig.~\ref{fig:analysis} (center) shows a strong agreement, with median $\rho=+0.641$,
$P(\rho > 0)=0.96$, and $P(\rho > 0.2)=0.91$. The regions where the critic predicts higher error are
essentially the regions where the flow matching representation moves the most between two probe layers, indicating that the critic's prediction shapes the actual generation update.

\paragraph{Experiment C: generation shift by critic's causal intervention.}
We finally test whether the alignment observed in Experiment~B is genuinely causal. To this end, we run inference twice on the same input: once without masking the critic token ($\mathrm{un}$), and once with the critic token masked out ($\mathrm{ma}$). Let $m_{\mathrm{un}}^{(\ell)}$ and $m_{\mathrm{ma}}^{(\ell)}$ denote the generation error $m^{(\ell)}$ (as defined in Experiment~B) under the unmasked and masked settings, respectively. We then define
\begin{equation}
Y_C = \big| (m_{\mathrm{un}}^{(42)} - m_{\mathrm{un}}^{(36)}) - (m_{\mathrm{ma}}^{(42)} - m_{\mathrm{ma}}^{(36)}) \big|,
\end{equation}
which is used to measure, between the two layers, how much of the $L36\!\to\!L42$ shift in generation error can be attributed to access to the critic.
As shown in Fig.~\ref{fig:analysis} (right), we observe a clear positive correlation, with median $\rho=+0.309$, $P(\rho > 0)=0.77$, and $P(\rho > 0.2)=0.62$. This suggests that the critic plays a causal role in steering the generation process.

\section{Conclusion}
\label{sec:conclusion}

We introduced \textbf{Inline Critic}, a learnable token that critiques a frozen editor at intermediate layers and steers its hidden states to refine generation within the forward pass, trained by a three-stage curriculum. Empirically, Critic-v2 achieves $7.89$ Overall on GEdit-Bench, $81.92$ on KRIS-Bench, $37.8$ on RISEBench ($+18.4$ over its backbone), showing consistent improvements across general editing and reasoning-heavy settings. Our analysis further shows that the attention and generation updates align with the critic's prediction, supporting Inline Critic as an effective refinement mechanism.

{
\small
\bibliographystyle{plainnat}
\bibliography{reference}

@article{li2025reflectdit,
  title={Reflect-{DiT}: Inference-Time Scaling for Text-to-Image Diffusion Transformers via In-Context Reflection},
  author={Li, Shufan and Kallidromitis, Konstantinos and Gokul, Akash and Koneru, Arsh and Kato, Yusuke and Kozuka, Kazuki and Grover, Aditya},
  journal={arXiv preprint arXiv:2503.12271},
  year={2025},
  url={https://arxiv.org/abs/2503.12271}
}

@inproceedings{zhuo2025reflectionflow,
  title={From Reflection to Perfection: Scaling Inference-Time Optimization for Text-to-Image Diffusion Models via Reflection Tuning},
  author={Zhuo, Le and Zhao, Liangbing and Paul, Sayak and Liao, Yue and Zhang, Renrui and Xin, Yi and Gao, Peng and Elhoseiny, Mohamed and Li, Hongsheng},
  booktitle={Proceedings of the IEEE/CVF International Conference on Computer Vision (ICCV)},
  pages={15329--15339},
  month={October},
  year={2025},
  url={https://openaccess.thecvf.com/content/ICCV2025/html/Zhuo_From_Reflection_to_Perfection_Scaling_Inference-Time_Optimization_for_Text-to-Image_Diffusion_ICCV_2025_paper.html}
}

@article{yin2025reasonedit,
  title={{ReasonEdit}: Towards Reasoning-Enhanced Image Editing Models},
  author={Yin, Fukun and Liu, Shiyu and Han, Yucheng and Wang, Zhibo and Xing, Peng and Wang, Rui and Cheng, Wei and Wang, Yingming and Li, Aojie and Yin, Zixin and Chen, Pengtao and Zhang, Xiangyu and Jiang, Daxin and Zeng, Xianfang and Yu, Gang},
  journal={arXiv preprint arXiv:2511.22625},
  year={2025},
  url={https://arxiv.org/abs/2511.22625}
}

@article{li2026thinkrledit,
  title={{ThinkRL-Edit}: Thinking in Reinforcement Learning for Reasoning-Centric Image Editing},
  author={Li, Hengjia and Jiang, Liming and Yan, Qing and Song, Yizhi and Kang, Hao and Liu, Zichuan and Lu, Xin and Wu, Boxi and Cai, Deng},
  journal={arXiv preprint arXiv:2601.03467},
  year={2026},
  url={https://arxiv.org/abs/2601.03467}
}

@article{li2025uniworldv2,
  title={{Uniworld-V2}: Reinforce Image Editing with Diffusion Negative-aware Finetuning and {MLLM} Implicit Feedback},
  author={Li, Zongjian and Liu, Zheyuan and Zhang, Qihui and Lin, Bin and Wu, Feize and Yuan, Shenghai and Yan, Zhiyuan and Ye, Yang and Yu, Wangbo and Niu, Yuwei and Wang, Shaodong and Cheng, Xinhua and Yuan, Li},
  journal={arXiv preprint arXiv:2510.16888},
  year={2025},
  url={https://arxiv.org/abs/2510.16888}
}

@inproceedings{luo2026editscore,
  title={{EditScore}: Unlocking Online {RL} for Image Editing via High-Fidelity Reward Modeling},
  author={Luo, Xin and Wang, Jiahao and Wu, Chenyuan and Xiao, Shitao and Jiang, Xiyan and Lian, Defu and Zhang, Jiajun and Liu, Dong and Liu, Zheng},
  booktitle={International Conference on Learning Representations (ICLR)},
  year={2026},
  url={https://openreview.net/forum?id=E7YpL4L4Xh}
}

@inproceedings{epstein2023selfguidance,
  title={Diffusion Self-Guidance for Controllable Image Generation},
  author={Epstein, Dave and Jabri, Allan and Poole, Ben and Efros, Alexei and Holynski, Aleksander},
  booktitle={Advances in Neural Information Processing Systems (NeurIPS)},
  volume={36},
  pages={16222--16239},
  publisher={Curran Associates, Inc.},
  year={2023},
  url={https://papers.nips.cc/paper_files/paper/2023/hash/3469b211b829b39d2b0cfd3b880a869c-Abstract-Conference.html}
}

@inproceedings{ahn2024pag,
  title={Self-Rectifying Diffusion Sampling with Perturbed-Attention Guidance},
  author={Ahn, Donghoon and Cho, Hyoungwon and Min, Jaewon and Jang, Wooseok and Kim, Jungwoo and Kim, SeonHwa and Park, Hyun Hee and Jin, Kyong Hwan and Kim, Seungryong},
  booktitle={Computer Vision -- ECCV 2024},
  pages={1--17},
  year={2024},
  doi={10.1007/978-3-031-73464-9_1},
  url={https://www.ecva.net/papers/eccv_2024/papers_ECCV/papers/09184.pdf}
}

@inproceedings{karras2024autoguidance,
  title={Guiding a Diffusion Model with a Bad Version of Itself},
  author={Karras, Tero and Aittala, Miika and Kynk{\"a}{\"a}nniemi, Tuomas and Lehtinen, Jaakko and Aila, Timo and Laine, Samuli},
  booktitle={Advances in Neural Information Processing Systems (NeurIPS)},
  volume={37},
  pages={52996--53021},
  publisher={Curran Associates, Inc.},
  year={2024},
  doi={10.52202/079017-1679},
  url={https://proceedings.neurips.cc/paper_files/paper/2024/hash/5ee7ed60a7e8169012224dec5fe0d27f-Abstract-Conference.html}
}

@inproceedings{koulischer2025feedback,
  title={Feedback Guidance of Diffusion Models},
  author={Koulischer, Felix and Handke, Florian and Deleu, Johannes and Demeester, Thomas and Ambrogioni, Luca},
  booktitle={Advances in Neural Information Processing Systems (NeurIPS)},
  year={2025},
  url={https://openreview.net/forum?id=8ySOcf7UpM}
}

@article{chefer2023attend,
  title={Attend-and-Excite: Attention-Based Semantic Guidance for Text-to-Image Diffusion Models},
  author={Chefer, Hila and Alaluf, Yuval and Vinker, Yael and Wolf, Lior and Cohen-Or, Daniel},
  journal={ACM Transactions on Graphics (SIGGRAPH)},
  volume={42},
  number={4},
  pages={1--10},
  year={2023},
  doi={10.1145/3592116},
  url={https://doi.org/10.1145/3592116}
}

@inproceedings{schusterbauer2026patchforcing,
  title={Denoising, Fast and Slow: Difficulty-Aware Adaptive Sampling for Image Generation},
  author={Schusterbauer, Johannes and Gui, Ming and Li, Yusong and Ma, Pingchuan and Krause, Felix and Ommer, Bj{\"o}rn},
  booktitle={IEEE/CVF Conference on Computer Vision and Pattern Recognition (CVPR)},
  year={2026},
  url={https://arxiv.org/abs/2604.19141}
}

@inproceedings{kou2024bayesdiff,
  title={{BayesDiff}: Estimating Pixel-wise Uncertainty in Diffusion via Bayesian Inference},
  author={Kou, Siqi and Gan, Lei and Wang, Dequan and Li, Chongxuan and Deng, Zhijie},
  booktitle={International Conference on Learning Representations (ICLR)},
  year={2024},
  url={https://openreview.net/forum?id=YcM6ofShwY}
}

@inproceedings{devita2025pixelwise,
  title={Diffusion Model Guided Sampling with Pixel-Wise Aleatoric Uncertainty Estimation},
  author={De Vita, Michele and Belagiannis, Vasileios},
  booktitle={Proceedings of the Winter Conference on Applications of Computer Vision (WACV)},
  pages={3844--3854},
  month={February},
  year={2025},
  url={https://openaccess.thecvf.com/content/WACV2025/html/De_Vita_Diffusion_Model_Guided_Sampling_with_Pixel-Wise_Aleatoric_Uncertainty_Estimation_WACV_2025_paper.html}
}

@inproceedings{lezama2022tokencritic,
  title={Improved Masked Image Generation with Token-Critic},
  author={Lezama, Jos{\'e} and Chang, Huiwen and Jiang, Lu and Essa, Irfan},
  booktitle={Computer Vision -- ECCV 2022},
  pages={70--86},
  year={2022},
  doi={10.1007/978-3-031-20050-2_5},
  url={https://www.ecva.net/papers/eccv_2022/papers_ECCV/html/2901_ECCV_2022_paper.php}
}

@inproceedings{chang2022maskgit,
  title={{MaskGIT}: Masked Generative Image Transformer},
  author={Chang, Huiwen and Zhang, Han and Jiang, Lu and Liu, Ce and Freeman, William T.},
  booktitle={Proceedings of the IEEE/CVF Conference on Computer Vision and Pattern Recognition (CVPR)},
  pages={11315--11325},
  month={June},
  year={2022},
  url={https://openaccess.thecvf.com/content/CVPR2022/html/Chang_MaskGIT_Masked_Generative_Image_Transformer_CVPR_2022_paper.html}
}

@inproceedings{ye2025imgedit,
  title={{ImgEdit}: A Unified Image Editing Dataset and Benchmark},
  author={Ye, Yang and He, Xianyi and Li, Zongjian and Lin, Bin and Yuan, Shenghai and Yan, Zhiyuan and Hou, Bohan and Yuan, Li},
  booktitle={Advances in Neural Information Processing Systems (NeurIPS) Datasets and Benchmarks Track},
  year={2025},
  url={https://openreview.net/forum?id=uUCSrMlfD3}
}

@article{liu2025step1xedit,
  title={{Step1X-Edit}: A Practical Framework for General Image Editing},
  author={Liu, Shiyu and Han, Yucheng and Xing, Peng and Yin, Fukun and Wang, Rui and Cheng, Wei and Liao, Jiaqi and Wang, Yingming and Fu, Honghao and Han, Chunrui and Li, Guopeng and Peng, Yuang and Sun, Quan and Wu, Jingwei and Cai, Yan and Ge, Zheng and Ming, Ranchen and Xia, Lei and Zeng, Xianfang and Zhu, Yibo and Jiao, Binxing and Zhang, Xiangyu and Yu, Gang and Jiang, Daxin},
  journal={arXiv preprint arXiv:2504.17761},
  year={2025},
  url={https://arxiv.org/abs/2504.17761}
}

@inproceedings{wu2025krisbench,
  title={{KRIS-Bench}: Benchmarking Next-Level Intelligent Image Editing Models},
  author={Wu, Yongliang and Li, Zonghui and Hu, Xinting and Ye, Xinyu and Zeng, Xianfang and Yu, Gang and Zhu, Wenbo and Schiele, Bernt and Yang, Ming-Hsuan and Yang, Xu},
  booktitle={Advances in Neural Information Processing Systems (NeurIPS) Datasets and Benchmarks Track},
  year={2025},
  url={https://openreview.net/forum?id=aWSh1Ec64T}
}

@inproceedings{zhao2025risebench,
  title={Envisioning Beyond the Pixels: Benchmarking Reasoning-Informed Visual Editing},
  author={Zhao, Xiangyu and Zhang, Peiyuan and Tang, Kexian and Zhu, Xiaorong and Li, Hao and Chai, Wenhao and Zhang, Zicheng and Xia, Renqiu and Zhai, Guangtao and Yan, Junchi and Yang, Hua and Yang, Xue and Duan, Haodong},
  booktitle={Advances in Neural Information Processing Systems (NeurIPS) Datasets and Benchmarks Track},
  year={2025},
  url={https://openreview.net/forum?id=T3UksaPK64}
}

@inproceedings{esser2024sd3,
  title={Scaling Rectified Flow Transformers for High-Resolution Image Synthesis},
  author={Esser, Patrick and Kulal, Sumith and Blattmann, Andreas and Entezari, Rahim and M{\"u}ller, Jonas and Saini, Harry and Levi, Yam and Lorenz, Dominik and Sauer, Axel and Boesel, Frederic and Podell, Dustin and Dockhorn, Tim and English, Zion and Rombach, Robin},
  booktitle={Proceedings of the 41st International Conference on Machine Learning},
  series={Proceedings of Machine Learning Research},
  volume={235},
  pages={12606--12633},
  publisher={PMLR},
  month={21--27 Jul},
  year={2024},
  url={https://proceedings.mlr.press/v235/esser24a.html}
}

@inproceedings{liu2023flow,
  title={Flow Straight and Fast: Learning to Generate and Transfer Data with Rectified Flow},
  author={Liu, Xingchao and Gong, Chengyue and Liu, Qiang},
  booktitle={International Conference on Learning Representations (ICLR)},
  year={2023},
  url={https://openreview.net/forum?id=XVjTT1nw5z}
}

@inproceedings{lipman2023flow,
  title={Flow Matching for Generative Modeling},
  author={Lipman, Yaron and Chen, Ricky T. Q. and Ben-Hamu, Heli and Nickel, Maximilian and Le, Matthew},
  booktitle={International Conference on Learning Representations (ICLR)},
  year={2023},
  url={https://openreview.net/forum?id=PqvMRDCJT9t}
}

@article{wan2025,
  title={{Wan}: Open and Advanced Large-Scale Video Generative Models},
  author={{Team Wan} and Wang, Ang and Ai, Baole and Wen, Bin and Mao, Chaojie and Xie, Chen-Wei and Chen, Di and Yu, Feiwu and Zhao, Haiming and Yang, Jianxiao and Zeng, Jianyuan and Wang, Jiayu and Zhang, Jingfeng and Zhou, Jingren and Wang, Jinkai and Chen, Jixuan and Zhu, Kai and Zhao, Kang and Yan, Keyu and Huang, Lianghua and Feng, Mengyang and Zhang, Ningyi and Li, Pandeng and Wu, Pingyu and Chu, Ruihang and Feng, Ruili and Zhang, Shiwei and Sun, Siyang and Fang, Tao and Wang, Tianxing and Gui, Tianyi and Weng, Tingyu and Shen, Tong and Lin, Wei and Wang, Wei and Wang, Wei and Zhou, Wenmeng and Wang, Wente and Shen, Wenting and Yu, Wenyuan and Shi, Xianzhong and Huang, Xiaoming and Xu, Xin and Kou, Yan and Lv, Yangyu and Li, Yifei and Liu, Yijing and Wang, Yiming and Zhang, Yingya and Huang, Yitong and Li, Yong and Wu, You and Liu, Yu and Pan, Yulin and Zheng, Yun and Hong, Yuntao and Shi, Yupeng and Feng, Yutong and Jiang, Zeyinzi and Han, Zhen and Wu, Zhi-Fan and Liu, Ziyu},
  journal={arXiv preprint arXiv:2503.20314},
  year={2025},
  url={https://arxiv.org/abs/2503.20314}
}

@article{wu2025qwenimage,
  title={{Qwen-Image} Technical Report},
  author={Wu, Chenfei and Li, Jiahao and Zhou, Jingren and Lin, Junyang and Gao, Kaiyuan and Yan, Kun and Yin, Sheng-ming and Bai, Shuai and Xu, Xiao and Chen, Yilei and Chen, Yuxiang and Tang, Zecheng and Zhang, Zekai and Wang, Zhengyi and Yang, An and Yu, Bowen and Cheng, Chen and Liu, Dayiheng and Li, Deqing and Zhang, Hang and Meng, Hao and Wei, Hu and Ni, Jingyuan and Chen, Kai and Cao, Kuan and Peng, Liang and Qu, Lin and Wu, Minggang and Wang, Peng and Yu, Shuting and Wen, Tingkun and Feng, Wensen and Xu, Xiaoxiao and Wang, Yi and Zhang, Yichang and Zhu, Yongqiang and Wu, Yujia and Cai, Yuxuan and Liu, Zenan},
  journal={arXiv preprint arXiv:2508.02324},
  year={2025},
  url={https://arxiv.org/abs/2508.02324}
}

@article{bai2025qwen25vl,
  title={{Qwen2.5-VL} Technical Report},
  author={Bai, Shuai and Chen, Keqin and Liu, Xuejing and Wang, Jialin and Ge, Wenbin and Song, Sibo and Dang, Kai and Wang, Peng and Wang, Shijie and Tang, Jun and Zhong, Humen and Zhu, Yuanzhi and Yang, Mingkun and Li, Zhaohai and Wan, Jianqiang and Wang, Pengfei and Ding, Wei and Fu, Zheren and Xu, Yiheng and Ye, Jiabo and Zhang, Xi and Xie, Tianbao and Cheng, Zesen and Zhang, Hang and Yang, Zhibo and Xu, Haiyang and Lin, Junyang},
  journal={arXiv preprint arXiv:2502.13923},
  year={2025},
  url={https://arxiv.org/abs/2502.13923}
}

@inproceedings{gal2023textualinversion,
  title={An Image is Worth One Word: Personalizing Text-to-Image Generation using Textual Inversion},
  author={Gal, Rinon and Alaluf, Yuval and Atzmon, Yuval and Patashnik, Or and Bermano, Amit Haim and Chechik, Gal and Cohen-Or, Daniel},
  booktitle={International Conference on Learning Representations (ICLR)},
  year={2023},
  url={https://openreview.net/forum?id=NAQvF08TcyG}
}

@article{dong2022dreamartist,
  title={{DreamArtist}++: Controllable One-Shot Text-to-Image Generation via Positive-Negative Adapter},
  author={Dong, Ziyi and Wei, Pengxu and Lin, Liang},
  journal={arXiv preprint arXiv:2211.11337},
  year={2022},
  url={https://arxiv.org/abs/2211.11337}
}

@inproceedings{huang2024reversion,
  title={{ReVersion}: Diffusion-Based Relation Inversion from Images},
  author={Huang, Ziqi and Wu, Tianxing and Jiang, Yuming and Chan, Kelvin C. K. and Liu, Ziwei},
  booktitle={SIGGRAPH Asia 2024 Conference Papers},
  pages={1--11},
  year={2024},
  doi={10.1145/3680528.3687658},
  url={https://ziqihuangg.github.io/projects/reversion}
}

@article{alaluf2023neti,
  title={A Neural Space-Time Representation for Text-to-Image Personalization},
  author={Alaluf, Yuval and Richardson, Elad and Metzer, Gal and Cohen-Or, Daniel},
  journal={ACM Transactions on Graphics (SIGGRAPH Asia)},
  volume={42},
  number={6},
  pages={243:1--243:10},
  year={2023},
  doi={10.1145/3618322},
  url={https://doi.org/10.1145/3618322}
}

@article{voynov2023pplus,
  title={{P+}: Extended Textual Conditioning in Text-to-Image Generation},
  author={Voynov, Andrey and Chu, Qinghao and Cohen-Or, Daniel and Aberman, Kfir},
  journal={arXiv preprint arXiv:2303.09522},
  year={2023},
  url={https://arxiv.org/abs/2303.09522}
}

@article{zhang2023prospect,
  title={{ProSpect}: Prompt Spectrum for Attribute-Aware Personalization of Diffusion Models},
  author={Zhang, Yuxin and Dong, Weiming and Tang, Fan and Huang, Nisha and Huang, Haibin and Ma, Chongyang and Lee, Tong-Yee and Deussen, Oliver and Xu, Changsheng},
  journal={ACM Transactions on Graphics (SIGGRAPH Asia)},
  volume={42},
  number={6},
  pages={244:1--244:14},
  year={2023},
  doi={10.1145/3618342},
  url={https://doi.org/10.1145/3618342}
}

@misc{blackforestlabs2025fluxkontext,
  title={{FLUX.1 Kontext}: Flow Matching for In-Context Image Generation and Editing in Latent Space},
  author={{Black Forest Labs} and Batifol, Stephen and Blattmann, Andreas and Boesel, Frederic and Consul, Saksham and Diagne, Cyril and Dockhorn, Tim and English, Jack and English, Zion and Esser, Patrick and Kulal, Sumith and Lacey, Kyle and Levi, Yam and Li, Cheng and Lorenz, Dominik and M{\"u}ller, Jonas and Podell, Dustin and Rombach, Robin and Saini, Harry and Sauer, Axel and Smith, Luke},
  year={2025},
  howpublished={Black Forest Labs technical report (arXiv:2506.15742)},
  url={https://arxiv.org/abs/2506.15742}
}

@article{deng2025bagel,
  title={Emerging Properties in Unified Multimodal Pretraining},
  author={Deng, Chaorui and Zhu, Deyao and Li, Kunchang and Gou, Chenhui and Li, Feng and Wang, Zeyu and Zhong, Shu and Yu, Weihao and Nie, Xiaonan and Song, Ziang and Shi, Guang and Fan, Haoqi},
  journal={arXiv preprint arXiv:2505.14683},
  year={2025},
  url={https://arxiv.org/abs/2505.14683}
}

@inproceedings{qin2025unicot,
  title={{Uni-CoT}: Towards Unified Chain-of-Thought Reasoning Across Text and Vision},
  author={Qin, Luozheng and Gong, Jia and Sun, Yuqing and Li, Tianjiao and Pan, Haoyu and Yang, Mengping and Yang, Xiaomeng and Qu, Chao and Tan, Zhiyu and Li, Hao},
  booktitle={International Conference on Learning Representations (ICLR)},
  year={2026},
  url={https://openreview.net/forum?id=5nevWRoNjn}
}

@misc{openai2025gpt4o,
  title={Introducing 4o Image Generation},
  author={{OpenAI}},
  year={2025},
  howpublished={OpenAI product release},
  url={https://openai.com/index/introducing-4o-image-generation/}
}

@misc{openai2025gptimage1,
  title={{GPT Image 1}},
  author={{OpenAI}},
  year={2025},
  howpublished={OpenAI API model card},
  url={https://platform.openai.com/docs/models/gpt-image-1}
}

@misc{openai2025gptimage15,
  title={{GPT Image 1.5}},
  author={{OpenAI}},
  year={2025},
  howpublished={OpenAI API model card},
  url={https://platform.openai.com/docs/models/gpt-image-1.5}
}

@misc{openai2026gptimage2,
  title={{GPT Image 2}},
  author={{OpenAI}},
  year={2026},
  howpublished={OpenAI API model card},
  url={https://platform.openai.com/docs/models/gpt-image-2}
}

@misc{google2025gemini2flash,
  title={Experiment with {Gemini 2.0 Flash} native image generation},
  author={Kampf, Kat and Brichtova, Nicole},
  year={2025},
  howpublished={Google Developers Blog},
  url={https://developers.googleblog.com/en/experiment-with-gemini-20-flash-native-image-generation}
}

@misc{google2025gemini25flashimage,
  title={Image editing in {Gemini} just got a major upgrade},
  author={Sharon, David and Brichtova, Nicole},
  year={2025},
  howpublished={Google Blog; introduces {Gemini 2.5 Flash Image} / Nano Banana},
  url={https://blog.google/products/gemini/updated-image-editing-model}
}

@misc{google2025gemini3proimage,
  title={Introducing Nano Banana Pro},
  author={Raisinghani, Naina},
  year={2025},
  howpublished={Google Blog; introduces {Gemini 3 Pro Image}},
  url={https://blog.google/innovation-and-ai/products/nano-banana-pro}
}

@article{bytedance2025doubao,
  title={{SeedEdit 3.0}: Fast and High-Quality Generative Image Editing},
  author={Wang, Peng and Shi, Yichun and Lian, Xiaochen and Zhai, Zhonghua and Xia, Xin and Xiao, Xuefeng and Huang, Weilin and Yang, Jianchao},
  journal={arXiv preprint arXiv:2506.05083},
  year={2025},
  url={https://arxiv.org/abs/2506.05083}
}

@inproceedings{wei2025omniedit,
  title={{OmniEdit}: Building Image Editing Generalist Models Through Specialist Supervision},
  author={Wei, Cong and Xiong, Zheyang and Ren, Weiming and Du, Xeron and Zhang, Ge and Chen, Wenhu},
  booktitle={International Conference on Learning Representations (ICLR)},
  year={2025},
  url={https://openreview.net/forum?id=Hlm0cga0sv}
}

@inproceedings{hui2025hqedit,
  title={{HQ-Edit}: A High-Quality Dataset for Instruction-based Image Editing},
  author={Hui, Mude and Yang, Siwei and Zhao, Bingchen and Shi, Yichun and Wang, Heng and Wang, Peng and Xie, Cihang and Zhou, Yuyin},
  booktitle={International Conference on Learning Representations (ICLR)},
  year={2025},
  url={https://openreview.net/forum?id=mZptYYttFj}
}

@inproceedings{zhao2024ultraedit,
  title={{UltraEdit}: Instruction-based Fine-Grained Image Editing at Scale},
  author={Zhao, Haozhe and Ma, Xiaojian and Chen, Liang and Si, Shuzheng and Wu, Rujie and An, Kaikai and Yu, Peiyu and Zhang, Minjia and Li, Qing and Chang, Baobao},
  booktitle={Advances in Neural Information Processing Systems (NeurIPS) Datasets and Benchmarks Track},
  year={2024},
  url={https://openreview.net/forum?id=9ZDdlgH6O8}
}

@article{chen2025sharegpt4oimage,
  title={{ShareGPT-4o-Image}: Aligning Multimodal Models with {GPT-4o}-Level Image Generation},
  author={Chen, Junying and Cai, Zhenyang and Chen, Pengcheng and Chen, Shunian and Ji, Ke and Wang, Xidong and Yang, Yunjin and Wang, Benyou},
  journal={arXiv preprint arXiv:2506.18095},
  year={2025},
  url={https://arxiv.org/abs/2506.18095}
}

@article{wind2025opengpt4o,
  title={{OpenGPT-4o-Image}: A Comprehensive Dataset for Advanced Image Generation and Editing},
  author={Chen, Zhihong and Bai, Xuehai and Shi, Yang and Fu, Chaoyou and Zhang, Huanyu and Wang, Haotian and Sun, Xiaoyan and Zhang, Zhang and Wang, Liang and Zhang, Yuanxing and Wan, Pengfei and Zhang, Yi-Fan},
  journal={arXiv preprint arXiv:2509.24900},
  year={2025},
  url={https://arxiv.org/abs/2509.24900}
}

@article{maplebb2025uniredit,
  title={{UniREditBench}: A Unified Reasoning-based Image Editing Benchmark},
  author={Han, Feng and Wang, Yibin and Li, Chenglin and Liang, Zheming and Wang, Dianyi and Jiao, Yang and Wei, Zhipeng and Gong, Chao and Jin, Cheng and Chen, Jingjing and Wang, Jiaqi},
  journal={arXiv preprint arXiv:2511.01295},
  year={2025},
  url={https://arxiv.org/abs/2511.01295}
}

@article{apple2025picobanana,
  title={{Pico-Banana-400K}: A Large-Scale Dataset for Text-Guided Image Editing},
  author={Qian, Yusu and Bocek-Rivele, Eli and Song, Liangchen and Tong, Jialing and Yang, Yinfei and Lu, Jiasen and Hu, Wenze and Gan, Zhe},
  journal={arXiv preprint arXiv:2510.19808},
  year={2025},
  url={https://arxiv.org/abs/2510.19808}
}

@misc{yejy2025nanoconsistent,
  title={{Nano-consistent-150k}},
  author={{Yejy53}},
  year={2025},
  howpublished={Hugging Face dataset},
  url={https://huggingface.co/datasets/Yejy53/Nano-consistent-150k}
}

@misc{qwen2026qwen35,
  title={{Qwen3.5-27B}},
  author={{Qwen Team}},
  year={2026},
  howpublished={Hugging Face model card},
  url={https://huggingface.co/Qwen/Qwen3.5-27B}
}

@article{mnih2015dqn,
  title={Human-level control through deep reinforcement learning},
  author={Mnih, Volodymyr and Kavukcuoglu, Koray and Silver, David and Rusu, Andrei A. and Veness, Joel and Bellemare, Marc G. and Graves, Alex and Riedmiller, Martin and Fidjeland, Andreas K. and Ostrovski, Georg and Petersen, Stig and Beattie, Charles and Sadik, Amir and Antonoglou, Ioannis and King, Helen and Kumaran, Dharshan and Wierstra, Daan and Legg, Shane and Hassabis, Demis},
  journal={Nature},
  volume={518},
  number={7540},
  pages={529--533},
  month={February},
  year={2015},
  doi={10.1038/nature14236},
  url={https://doi.org/10.1038/nature14236}
}
}

\newpage

%%%%%%%%%%%%%%%%%%%%%%%%%%%%%%%%%%%%%%%%%%%%%%%%%%%%%%%%%%%%

\appendix

\section*{Appendix}

\section{Details of Implementation}
\label{appendix:training_details}
\paragraph{Training.}
We build Inline Critic on two public Qwen-Image-Edit~\citep{wu2025qwenimage} backbones, Qwen-Image-Edit-2509 and Qwen-Image-Edit-2511, yielding two variants that we refer to as Critic-v1 and Critic-v2. We use $K=9$ probes at $\mathcal{L}^* = \{6, 12, 18, 24, 30, 36, 42, 48, 54\}$, with $256$-dim outputs for $\psi^{(\ell)}$ and $\chi^{(\ell)}$. The training set is a $2.2$M-pair mixture from OmniEdit~\citep{wei2025omniedit}, UltraEdit~\citep{zhao2024ultraedit}, HQ-Edit~\citep{hui2025hqedit}, ShareGPT-4o-Image~\citep{chen2025sharegpt4oimage}, OpenGPT-4o-Image~\citep{wind2025opengpt4o}, UniREdit-Data-100K~\citep{maplebb2025uniredit}, Nano-Consistent-150K~\citep{yejy2025nanoconsistent}, and the SFT split of Pico-Banana-400K~\citep{apple2025picobanana}, processed at $512\!\times\!512$. For classifier-free guidance, the instruction is dropped with probability $0.05$, and with probability $0.5$ replaced by an offline Qwen3.5-27B~\citep{qwen2026qwen35} rewrite using the ThinkRL-Edit~\citep{li2026thinkrledit} prompt. Each stage runs for $4$ epochs with AdamW (lr $1\!\times\!10^{-4}$, $\beta=(0.9,0.999)$, weight decay $0.01$) and a cosine-with-restarts schedule (min-LR ratio $0.1$, $500$ warmup steps). The per-device batch is $4$ with gradient accumulation $4$, on $4$ nodes of $8$ A100-80G GPUs. Stage~3 uses $\lambda_c=\lambda_p=1$ in Eq.~\ref{eq:stage3}.

\paragraph{Testing.}
We evaluate on RISEBench~\citep{zhao2025risebench}, KRIS-Bench~\citep{wu2025krisbench}, GEdit-Bench~\citep{liu2025step1xedit}, and ImgEdit~\citep{ye2025imgedit}, with the same Qwen3.5-27B rewrite applied on RISEBench and KRIS-Bench. We resize inputs to a $\sim$$1024^2$-pixel budget at the original aspect ratio, and use $40$ denoising steps at guidance scale $4.0$.

\section{Qualitative Results}
\label{app:qualitative}

\begin{figure*}[h]
\centering
\includegraphics[width=\textwidth]{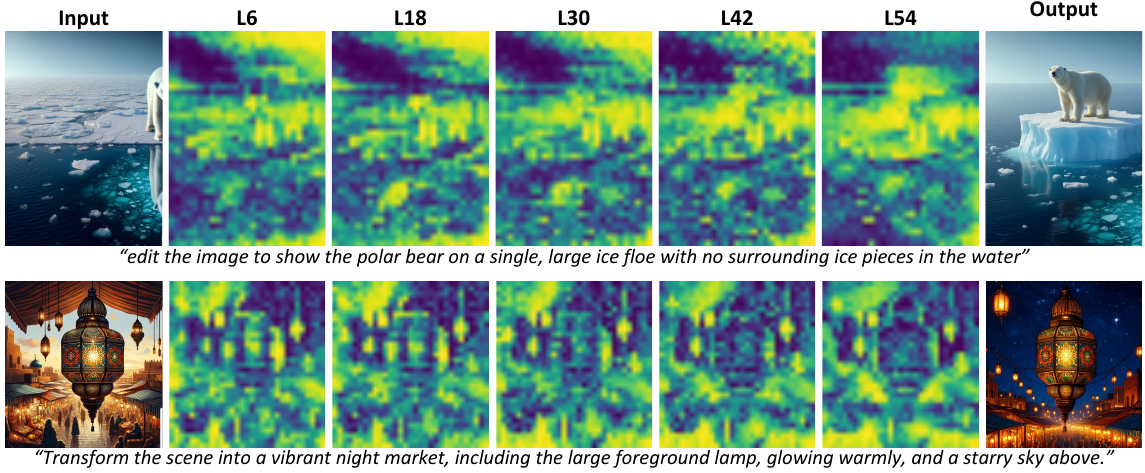}
\caption{\textbf{Qualitative results.} We visualize the probe-layer error maps across backbone layers
($L6\!\to\!L54$), together with the input image (left) and the final edited result (right).}
\label{fig:qualitative}
\end{figure*}

We provide qualitative results in Fig.~\ref{fig:qualitative}. Starting from the early backbone layers, the probe-layer error maps highlight broad regions that are likely to require modification. As the depth increases from $L6$ to $L54$, these regions gradually become more focused, indicating that the critic refines the model's internal behavior throughout the forward pass. The final outputs demonstrate that this progressive refinement leads to edits that are both instruction-faithful and visually coherent.

\section{Limitations}
\label{app:limitations}

The critic is trained to predict the per-position MSE of the probe head, but the training data can be highly noisy. Therefore it is not always a faithful measure of generation quality, which leaves the critic's learning signal correspondingly noisy. A natural next step is to replace the MSE target with a feature-space criterion, so that the critic learns perceptually meaningful errors.

%%%%%%%%%%%%%%%%%%%%%%%%%%%%%%%%%%%%%%%%%%%%%%%%%%%%%%%%%%%%

% \newpage
% \input{checklist.tex}

\end{document}